\documentclass[journal]{IEEEtran}

\usepackage{flushend}
\usepackage{graphicx}
\graphicspath{{images/}}
\usepackage{algorithm}
\usepackage{algorithmicx}
\usepackage{algpseudocode}
\usepackage{amsmath}
\usepackage{amssymb}
\usepackage{nicefrac}
\usepackage{amsfonts}
\usepackage{mathtools}
\usepackage{siunitx} 
\usepackage{textcomp}
\usepackage{algpseudocode}

\usepackage{comment}

\usepackage{nomencl}
\makenomenclature
\providecommand{\algorithmname}{Algorithm}
\floatname{algorithm}{\protect\algorithmname}
\algnewcommand{\LeftComment}[1]{\Statex \textit{#1}}
\algnewcommand\algorithmicinput{\textbf{Define:}}
\algnewcommand\DEFINE{\item[\algorithmicinput]}

\usepackage{tabularx,booktabs}
\usepackage{gensymb}
\usepackage{multirow}
\usepackage[caption=false,font=footnotesize]{subfig}
\usepackage{arydshln}

\usepackage[pagebackref=true,breaklinks=true,colorlinks=true,bookmarks=false]{hyperref}

\usepackage[noadjust]{cite}

\begin{document}

\title{Probabilistic Semantic Segmentation Refinement by Monte Carlo Region Growing}

\author{Philipe A. Dias,~\IEEEmembership{Student Member,~IEEE,}
          Henry Medeiros,~\IEEEmembership{Senior Member, IEEE}
\thanks{Department of Electrical and Computer Engineering, Marquette University, Milwaukee, WI, 53233, e-mail: philipe.ambroziodias@marquette.edu \hspace{2cm} henry.medeiros@marquette.edu.}
\thanks{We acknowledge the support of USDA ARS agreement \#584080-5-020, and of NVIDIA Corporation with the donation of the GPU used for this research.}
}

\markboth{Submitted to IEEE Transactions on Image Processing (under review), Apr~2020}%
{P. Dias and H. Medeiros: Tracking Passengers and Divested Items}

\maketitle


\IEEEtitleabstractindextext{%
}

\begin{abstract}
Semantic segmentation with fine-grained pixel-level accuracy is a fundamental component of a variety of computer vision applications. However, despite the large improvements provided by recent advances in the architectures of convolutional neural networks, segmentations provided by modern state-of-the-art methods still show limited boundary adherence. We introduce a fully unsupervised post-processing algorithm that exploits Monte Carlo sampling and pixel similarities to propagate high-confidence pixel labels into regions of low-confidence classification. Our algorithm, which we call \textit{probabilistic Region Growing Refinement} (pRGR), is based on a rigorous mathematical foundation in which clusters are modelled as multivariate normally distributed sets of pixels. Exploiting concepts of Bayesian estimation and variance reduction techniques, pRGR performs multiple refinement iterations at varied receptive fields sizes, while updating cluster statistics to adapt to local image features. Experiments using multiple modern semantic segmentation networks and benchmark datasets demonstrate the effectiveness of our approach for the refinement of segmentation predictions at different levels of coarseness, as well as the suitability of the variance estimates obtained in the Monte Carlo iterations as uncertainty measures that are highly correlated with segmentation accuracy. 
\end{abstract}
\section{Introduction}
\IEEEPARstart{F}{or} many applications of computer vision, image segmentation with high accuracy at pixel-level is a key requirement. In action and activity recognition, relevant visual cues for human-human and human-object interactions include contact between agent and object, particular body silhouettes, and part locations \cite{shao2012human,ma2016going,gupta2009observing}. Automation tasks often require manipulation of objects or instruments, where the quality of object pose and morphology estimation directly impact success rate \cite{wang2016robot,shvets2018automatic}. The agricultural field, where image segmentation has been exploited as part of perception modules targeting pollination, orchard management, and harvesting in horticultural scenarios  is an example \cite{Williams2020_kiwi,dias_multispecies_2018,Tabb2019_orchard}.

The wide range of image segmentation applications includes image editing, self-driving vehicles \cite{Janai17_vehicles}, virtual clothing try-on for online shopping \cite{han2018viton}, and medical imaging. This is exemplified by the Medical Segmentation Decathlon Challenge \cite{simpson2019_decathlon}, a scenario where precise localization of organs and structures such as tumors is crucial for eventual guidance of medical interventions.

Deep learning models based on convolutional neural networks (CNN) have substantially improved the state of the art in image understanding. However, conventional CNN-based segmentation models are limited by the typical downsampling employed to learn hierarchical features. Pixel-level details are lost in this process, resulting in segmentation masks that poorly adhere to object boundaries.

To mitigate these limitations, modern image segmentation models employ strategies such as \textit{atrous} convolutions \cite{Chen2014_Deeplab1}, encoder-decoder architectures with skip-connections \cite{Long2015,Ronneberger,Lin2016}, pyramid scaling \cite{Chen2015_Deeplab2}, among others. Large improvements have been achieved through these strategies in comparison to conventional CNN architectures, but the segmentation they produce still tends not to be finely aligned with the boundaries of objects. Post-processing approaches such as conditional random fields (CRFs) \cite{Chen2015_Deeplab2,Krahenbuhl2012} have been successful in segmentation refinement, but their performance depends on proper optimization of parameters for each specific dataset and predictor module being used.

In \cite{dias_rgr_2018}, we introduced the Region Growing Refinement (RGR) algorithm, an alternative unsupervised and easily generalizable post-processing module that refines semantic segmentation masks by means of appearance-based region growing. In a Monte Carlo framework, initial pixels are sampled as high-quality seeds from regions labeled with high-confidence scores and grown into clusters for segmentation refinement. In this context,  we present the \textit{probabilistic Region Growing Refinement (pRGR)} algorithm, an extension of RGR that provides the following contributions:
\begin{itemize}
    \item a solid mathematical foundation that exploits a probabilistic framework to guide all the steps of the algorithm;
    \item combining techniques from Bayesian estimation, many parameters that were previously determined in an ad-hoc manner are now initialized using Bayesian conjugate priors and updated as assignments of pixels to clusters occur. Moreover, variance reduction techniques are exploited to optimize the sampling steps within the Monte Carlo refinement iterations;
    \item with a novel parameterization that allows for the emulation of varied receptive field sizes, pRGR further improves segmentation refinement performance by recovering finer boundary details and attenuating the effects of false-positive pixel labels;
    \item we experimentally demonstrate the applicability of pRGR in a variety of scenarios that include state-of-the-art models such as DeepLabV3+ \cite{Chen2017_Deeplab3}. Such experiments also suggest the combination of DenseCRF \cite{Krahenbuhl2012} and pRGR as a powerful strategy for segmentation refinement;
    \item we observe that the variance of pRGR's Monte Carlo estimations can be exploited as an uncertainty estimation mechanism, with experiments demonstrating its high correlation with final segmentation accuracy values;
    \item upon publication, code will be made available at \url{coviss.org/code}.
\end{itemize}

We report experiments using different CNNs, datasets and baselines. For easy comparison against CRF and RGR baselines, we first report experiments on refinement of segmentation predictions provided by DeepLab \cite{Chen2014_Deeplab1} and DeepLabV2 \cite{Chen2015_Deeplab2} for the PASCAL VOC 2012 \cite{pascal2012} validation set. We then report experiments conducted with the state-of-the-art DeepLabV3+ \cite{Chen2017_Deeplab3} segmentation model on the PASCAL \textit{val} set and also on selected sequences from the DAVIS dataset \cite{Perazzi2016}. Compared to the PASCAL dataset, the DAVIS dataset contains annotations that are more fine-grained, with tighter boundary adherence.

The paper is organized as follows. In Section \ref{sec:relworks}, we provide an overview of the related work,  which includes modern semantic segmentation models, segmentation refinement strategies and clustering algorithms that use similar probabilistic concepts. The complete formulation of our pRGR model is then explained in Section \ref{sec:method}, while Section \ref{sec:algorithm} details the algorithm that implements pRGR. In Section \ref{sec:experiments}, we report experiments where pRGR is compared with RGR, CRF and a combination of CRF+pRGR for refinement of predictions provided by multiple CNN models. Finally, in Section \ref{sec:conclusions} we stress the main takeaways of this work, both in terms of obtained results as well as future directions where pRGR can be exploited.

\section{Related Work}\label{sec:relworks}

Models based on deep CNNs have remarkably advanced the state of the art in most computer vision tasks, such as image classification and object detection. Yet, tasks requiring image labeling at pixel-level are particularly challenging for CNN-based systems. While crucial for evaluating varied levels of context and thus learning hierarchical features, the combination of \textit{pooling} and striding operations leads to a downsampling effect that compromises the performance of CNNs for pixel-dense classification tasks. This is clearly exemplified by segmentation predictions generated by models such as the ones introduced by Eigen \& Fergus \cite{eigen2015predicting} and the earlier Fully Convolutional Networks (FCNs) \cite{Long2015}, whose architectures essentially consisted of image classification CNN models with their fully connected layers replaced by further convolutions. These models generate coarse segmentation masks with limited boundary adherence, an open problem that has driven many advances in the field. 

Many current approaches for semantic segmentation focus on developing better upsampling strategies to improve segmentation accuracy. Noh et al. \cite{Noh2015} focused on learning a deconvolution network, while works such as U-Net \cite{Ronneberger} and SegNet \cite{Badrinarayanan2015segnet} focused on encoder-decoder architectures where the decoder path includes skip-connections to convey information from encoder layers to better guide upsampling. 

Another direction that has been investigated to obtain finer segmentations concentrates on reducing the amount of details lost through downsampling. To that end, the family of DeepLab models \cite{Chen2014_Deeplab1,Chen2015_Deeplab2,Chen2017_Deeplab3} exploits the idea of dilated (or \textit{atrous}) convolutions, where convolutional filters are padded with zeros as an alternative way to increase receptive fields.

Moreover, works such as PSPNet \cite{zhao2017pyramid} revisit earlier strategies \cite{grauman2005pyramid} that focus on evaluating images at multiple scales to better incorporate various levels of scene context. In this context, DeepLabV2 \cite{Chen2015_Deeplab2} employs atrous spatial pyramid pooling (ASPP), where atrous convolutions are combined with the concept of Spatial Pyramid Pooling \cite{he2015spatial}. More recently, the current state-of-the-art DeepLabV3+ model \cite{Chen2017_Deeplab3} was introduced, combining ASPP strategies adjusted to exploit image-level features and a decoder module to refine segmentation along boundaries. 

\begin{figure*}[t]
    \centering
    \includegraphics[width=\linewidth]{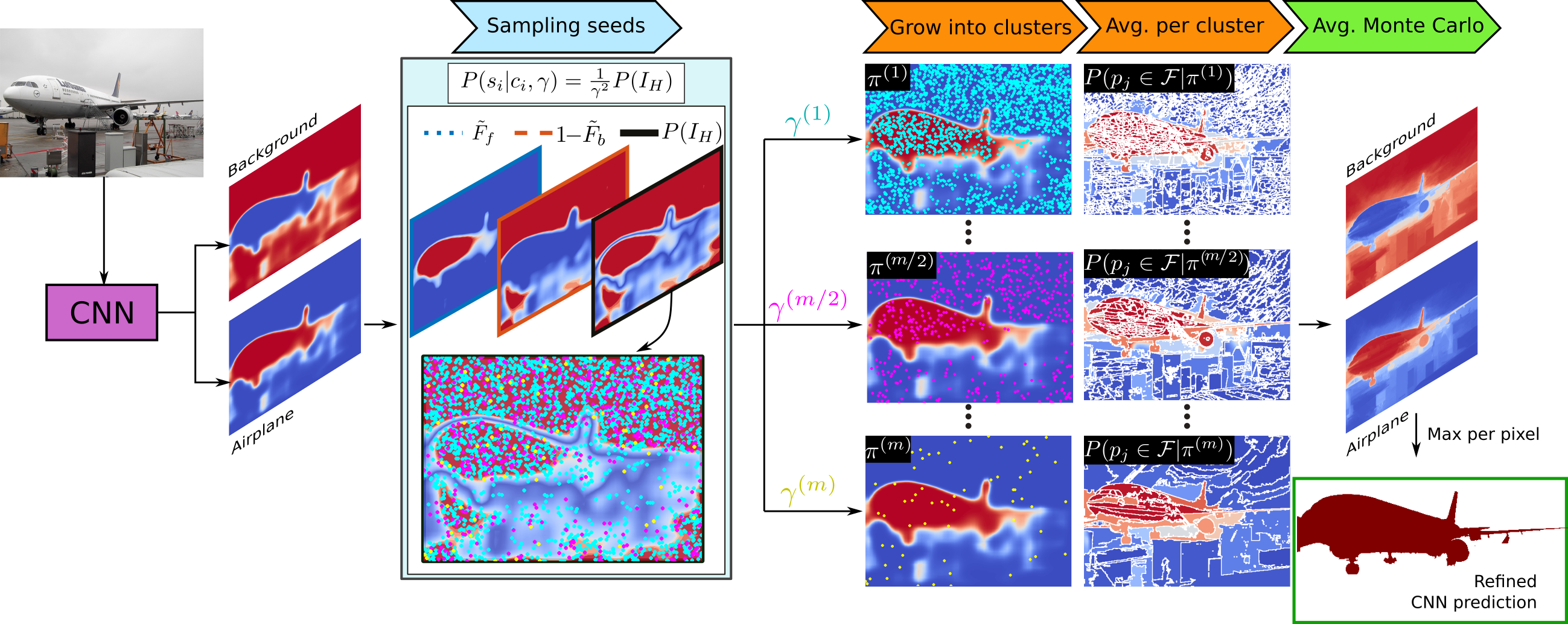}
    \caption{Diagram illustrating the sequence of steps performed by the proposed pRGR model for segmentation refinement. Each step and their corresponding equations are discussed in detail in Section \ref{sec:method}.}
    \label{fig:flow}
\end{figure*}

In addition to adjustments in CNN architectures, some studies focus on investigating techniques that employ low-level image features to aid CNN-based models in image segmentation tasks. The concept of Selective Search \cite{Uijlings2013} was exploited by Girschick et al. \cite{Girshick2014} to conceive the first model of the RCNN family for object detection. Sets of small regions \cite{Felzenszwalb} are merged based on similarity to generate region proposals which are then evaluated using a deep learning model. Similar ideas exploit superpixels \cite{Stutz} as a pre-processing step, where pixels are grouped based on low-level properties (e.g. color similarity) and each group is evaluated using hand-engineered hierarchical features \cite{Gould2008} or CNNs \cite{Couprie2013,Mostajabi}.

Likewise, local-appearance techniques such as superpixels and conditional random fields (CRFs) have also been employed for the post-processing of segmentations generated by deep CNN models. The DeepLab paper \cite{Chen2014_Deeplab1} proposes to integrate its novel architecture with the DenseCRF model from \cite{Krahenbuhl2012} to refine segmentation masks especially along boundaries. In contrast  to conventional fully connected CRFs implementations, DenseCRF improves computational efficiency thanks to an approximate inference algorithm in which pairwise potentials are modelled as combinations of Gaussian kernels. However, using the DenseCRF model for post-processing refinement of segmentation masks requires optimization of hyperparameters through grid-search, a process that must be performed whenever the CNN model and/or the dataset is changed. 

In \cite{dias_rgr_2018}, we introduced the Region Growing Refinement (RGR) algorithm, which refines segmentation predictions by propagating high-confidence labels into regions of uncertain pixel classification. Experiments on different combinations of datasets and CNN models demonstrated: i) RGR's efficacy for segmentation refinement; and ii) its high generalization capabilities, not requiring dataset- or model-specific adjustments. In addition to segmentation refinement \cite{dias_multispecies_2018}, the practical relevance of RGR is also illustrated in FreeLabel \cite{dias_freelabel_2018}, an open-source annotation tool where high-quality segmentation masks can be obtained from user-provided freehand traces.

While similar to superpixel algorithms such as SLIC \cite{Achanta2012} in some aspects, RGR's initialization of seeds based on random sampling from high-confidence regions allows for clusters of flexible sizes and enforces the classification of high-uncertainty regions to be derived from high-confidence ones. Additional limitations of conventional superpixel algorithms that are however shared by RGR include lack of adaptiveness in terms of adjusting to local features as well as poor robustness to mistakes in the initialization of parameters. 

Models exploiting Bayesian estimation have been introduced to overcome these limitations of superpixel algorithms, with strategies that range from pixel-related Gaussian Mixture Models (GMMs) \cite{freifeld2015sppx,ban2018superpixel} to non-parametric mixture models \cite{uziel2019bayesSppx}. In such approaches, previously fixed normalization hyperparameters are replaced by Bayesian priors, which are updated in conjunction with other cluster statistics in the form of covariances as assignments of pixels to clusters take place.

\section{Proposed Approach} \label{sec:method}
In this section, we first briefly review the main operations composing the RGR algorithm. Then, we describe the sequence of steps and corresponding mathematical formulation that comprise our probabilistic Region Growing Refinement (pRGR) method.

{\textbf{Region Growing Refinement (RGR):}} Based on pixel classification scores available from a semantic segmentation detector (e.g. a modern CNN), RGR identifies three regions in the image: high confidence background, high confidence object, and uncertainty region. This is performed by thresholding the scoremaps using extreme values, i.e., near $1.0$ for high-confidence foreground and near $0.0$ for high-confidence background identification. Region growing based on pixel color and location similarity is then performed, starting from initial seeds that are sampled from high-confidence regions. RGR performs this process multiple times using a Monte Carlo approach: different sets of seeds are randomly sampled for each growing iteration, such that the overall impact of eventually sampling false-positive pixels as seeds is minimized. Once the clusters are formed, RGR conducts a pixel-based majority voting within each cluster to obtain a refined estimate of the segmentation scores for each region. Finally, refined scores collected from each Monte Carlo iteration are averaged to obtain the resulting refined segmentation predictions.

Similar to RGR, \emph{the proposed pRGR algorithm is a generic unsupervised post-processing module} for refinement of segmentation boundaries that can be coupled with the output of any CNN or similar model for semantic segmentation. While sharing similar concepts, pRGR advances RGR by employing a probabilistic formulation in which all the steps of the algorithm are derived using a mathematically coherent framework. In addition, concepts of variance reduction and Bayesian estimation are used for the initialization and update of parameters in a principled manner.

The main operations composing pRGR are summarized in Fig. \ref{fig:flow}. At a high level, the steps performed by both RGR and pRGR can be summarized as: 1) identification of high confidence classification regions; 2) Monte Carlo seed sampling from high-confidence regions; 3) region growing of seeds into clusters; 4) pixel-score averaging within clusters; 5) averaging across multiple Monte Carlo iterations. In the case of multi-class segmentation, both RGR and pRGR perform these steps on the scoremaps associated with each class, and the final classification is defined by computing the maximum likelihood across classes. In the remainder of this section, we justify these operations and derive the set of equations guiding the steps composing our method. 

\begin{figure*}[t]
    \centering
    \includegraphics[height=4cm]{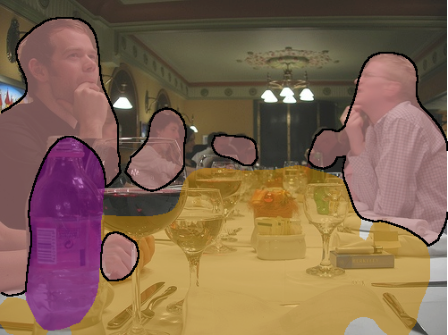}
    \includegraphics[height=4cm]{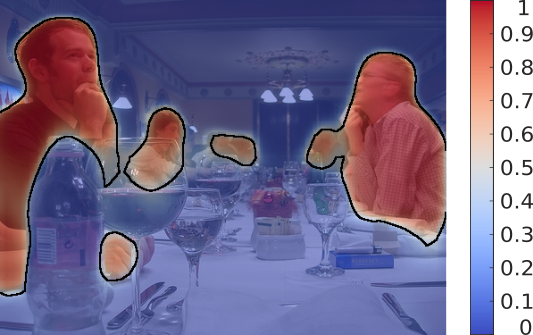}\hspace{10pt}
    \includegraphics[height=4cm]{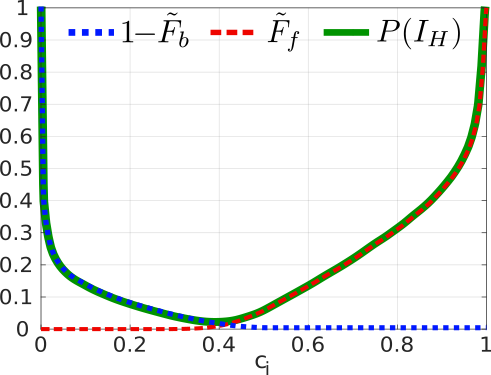}
    \caption{Example of non-parametric estimation of the probability $P(I_H)$ that a pixel is part of the region labeled with high-confidence. \textit{Left:} original segmentation collected from a CNN. \textit{Center:} pixel scores predicted by the CNN for the \textit{person} category. \textit{Right:} $\Tilde{F}_b,\Tilde{F}_f$ are the cumulative density functions of scores for non-person (i.e., ``background'') and person (foreground) pixels, respectively. }
    \label{fig:threshold_dist}
\end{figure*}

\subsection{Probabilistic seed sampling from high-confidence regions}
Let the inputs for our refinement algorithm be represented as an observed image $I\in\mathbb{R}^{w\times h}$ and corresponding confidence maps $C\in\mathbb{R}^{w\times h\times \mathcal{C}}$. Here, $w\times h$ are the dimensions of the input image $I$, and $C$ are the scoremaps for each class in the set $\mathcal{C}$, generated by any modern segmentation CNN. For simplicity, we first introduce the method for the binary case where $|\mathcal{C}|=1$, as all steps are performed on each class scoremap independently in the multiclass scenario. 

Let $\pi$ represent the partition of $I$ into a set of clusters $\pi=\left\{ \psi_{1},\psi_{2},\ldots,\psi_{|S|}\right\} $, which are grown from the set of seeds ${S}=\left\{ s_{1},s_{2},\ldots,s_{|S|}\right\} $. To estimate the probability that a pixel $p_i$ should be sampled as a high-confidence seed $s_i$, let the thresholds defining high-confidence background and high-confidence foreground be denoted by $t_{b}$ and $t_{f}$, respectively. From that, we define $I_{H}=\left\{ c_{i}< t_b\text{ or }c_{i}\geq t_f\right\} $ as the event that a pixel with confidence score $c_{i}$ belongs to a high-confidence background or foreground region. The probability $P(I_H)$ is thus given by 
\begin{align}
P(I_{H}) =& P(c_{i}<t_b\text{ or }c_{i}\geq t_f)\nonumber\\
 =&P(c_i<t_b)+P(c_i\geq t_f)-P(c_i<t_b)P(c_i\geq t_f)\nonumber\\
 =&1-P(c_i\geq t_b)+P(c_i\geq t_f)\nonumber\\&\hspace{7em}-[1-P(c_i\geq t_b)]P(c_i\geq t_f)\nonumber\\
  =&1-P(c_i\geq t_b)+P(c_i\geq t_b)P(c_i\geq t_f)\nonumber\\
 =& 1-F_b\left(c_{i}\right)+F_b\left(c_{i}\right)F_f\left(c_{i}\right)\label{eq:ptlth},
\end{align}
where $F_b(\cdot)$ and $F_f(\cdot)$ are the cumulative density functions (CDFs) corresponding to the distributions of $t_b$ and $t_f$, respectively.

As discussed in \cite{dias_rgr_2018}, sampling with a spacing $\gamma$ between seeds ensures the availability of paths for them to grow throughout the uncertainty region. That is, seeds are uniformly sampled among $\gamma\times\gamma$ points within the high-confidence region, such that, given the thresholds $t_f,t_b$ and the inter-seed spacing $\gamma$, the probability of sampling a seed $s_i$ at a pixel with confidence score $c_i$ is
\begin{equation}
P(s_{i}|c_{i}< t_b \text{ or }c_{i}\geq t_f,\gamma)=\frac{1}{\gamma^{2}}.\label{eq:psi}
\end{equation}

While in RGR the seed spacing $\gamma$ is fixed for all sample-grow iterations, for pRGR we adopt a strategy where $\gamma$ is itself sampled in a stratified manner from an uniform distribution $\gamma \sim  \mathcal{U}(\gamma_{l},\gamma_{h})$, where $\gamma_{l}$ and $\gamma_{h}$ are the minimum and the maximum spacing values. As indicated by (\ref{eq:psi}), the parameter $\gamma$ directly impacts the number of seeds to be sampled, which is inversely proportional to the expected sizes of the clusters to be formed through seed growing. Hence,  \emph{sampling $\gamma$ using a stratified approach allows for the emulation of the refinement process at multiple receptive field sizes}, a common practice exploited in many modern segmentation architectures \cite{zhao2017pyramid,Chen2017_Deeplab3}.

Since $t_{l}$ and $t_{h}$ are independent of $\gamma$, we have 
\begin{alignat}{1}
P(s_{i},I_{H}|\gamma)= & P(s_{i}|I_{H},\gamma)P(I_{H})\nonumber\\
= & \frac{1}{\gamma^{2}}\left[1-F_b\left(c_{i}\right)+F_b\left(c_{i}\right)F_f\left(c_{i}\right)\right].\label{eq:condgamma}
\end{alignat}
Marginalizing over the event $I_{H}$,
\begin{alignat}{1}
P(s_{i}|\gamma) & =P(s_{i},I_{H}|\gamma)+P(s_{i},\bar{I}_{H}|\gamma)\nonumber\\
 & =P(s_{i},I_{H}|\gamma),\label{eq:lhcondgamma2}
\end{alignat}
where the second equation is based on the fact that seeds are sampled only from the high-confidence region, i.e., $P(s_{i}|\bar{I}_{H},\gamma)=0$.

Let $m=1,...,n_s$ represent the index of a Monte Carlo growing iteration, such that $s_{i}^{(m)}$ represents the $i$-th seed in iteration $m$, and let $\gamma^{(m)}$ be the corresponding inter-seed spacing. Based on (\ref{eq:condgamma}) and (\ref{eq:lhcondgamma2}), the seed samples are distributed according to
\begin{equation}
s_{i}^{(m)}\sim \frac{1}{{(\gamma^{(m)})}^{2}}\left[1-F_b\left(c_{i}\right)+F_b\left(c_{i}\right)F_f\left(c_{i}\right)\right]. \label{eq:si}
\end{equation}

\hspace{0pt}

\textbf{Thresholds distribution:}\label{sub:thdist}
Semantic segmentation methods based on deep-learning models typically comprise three main steps. First, a CNN computes unbounded scoremaps with the activations of each pixel for each class. By applying a softmax function across all the classes for each pixel, these scoremaps are then normalized into the range $[0,1]$. Finally, class labels are assigned to each pixel through an $\arg\max$ operation across the normalized scoremaps.

Therefore, no single fixed threshold is applied to the class scoremaps for classification. Hence, to estimate the CDFs $F_b,F_f$ required in (\ref{eq:condgamma}), we approximate them using two non-parametric distributions $\Tilde{F_b}$ and $\Tilde{F_f}$. As depicted in Fig.  \ref{fig:threshold_dist}, from the output of the $\arg\max$ step we identify the pixels $p_f \in \mathcal{F}$ labeled as foreground and the pixels $p_b \in \mathcal{B}$ labeled as background. For a scenario of multiple classes such as the one illustrated in Fig. \ref{fig:threshold_dist}, foreground corresponds to pixels labeled as part of the category under evaluation (e.g. \textit{person}), while background corresponds to the union of all the remaining categories (i.e., \textit{non-person}). Then, we estimate the CDFs $\Tilde{F_f}\approx F(c_f)$ and $\Tilde{F_b}\approx F(c_b)$ of the scores $c_f$ and $c_b$ computed by the CNN for the pixels predicted within foreground $\mathcal{F}$ and background $\mathcal{B}$, respectively. To that end, we use a normal kernel function that is evaluated at equally-spaced points over the range $[0,1]$ of normalized scores predicted for each region. 

\subsection{Similarity measurement}

Once in possession of high-confidence seeds, pRGR proceeds to grow these initial pixels into clusters based on spatial and color similarity. Let each pixel $p_j$ be described by a 5D feature vector $\mathbf{z_j}=[\mathbf{x_j},\mathbf{c_j}]^T$, where $\mathbf{x_j}=[x_j,y_j]^T$ are its 2D spatial features and $\mathbf{c_j}=[l_j,a_j,b_j]^T$ its 3D color (CIELab) features. Similarly, let $\mathbf{x_k}, \mathbf{c_k}$ represent the features of the centroid of a cluster $\psi_k$. Then, following the formulation in \cite{dias_rgr_2018} (which is based on the SLIC superpixel algorithm \cite{Achanta2012}), the similarity between $p_j$ and a cluster $\psi_k$ is given by
\begin{equation}\label{eq:eucdist}
d\left(p_j,\psi_k\right)=\frac{\left\Vert \mathbf{x_{j}}-\mathbf{x_{k}}\right\Vert_2^2 }{\sigma_{s}}+\frac{\left\Vert \mathbf{c_{j}}-\mathbf{c_{k}}\right\Vert_2^2 }{\sigma_{m}}.
\end{equation}
Equation (\ref{eq:eucdist}) can be generalized to
\begin{flalign}\label{eq:eucdist_general}
d\left(p_j,\psi_k\right)&=\frac{1}{\sigma_{s}}(\mathbf{x_{j}}-\mathbf{x_{k}})^{T}(\mathbf{x_{j}}-\mathbf{x_{k}}) \notag \\
& \hspace{6em} +\frac{1}{\sigma_{m}}(\mathbf{c_{j}}-\mathbf{c_{k}})^{T}(\mathbf{c_{j}}-\mathbf{c_{k}})\nonumber \\
 &=(\mathbf{x_{j}}-\mathbf{x_{k}})^{T}\mathbf{\Sigma_{x}}^{-1}(\mathbf{x_{j}}-\mathbf{x_{k}}) \nonumber\\
 &\hspace{6em}+(\mathbf{c_{j}}-\mathbf{c_{k}})^{T}\mathbf{\Sigma_{c}}^{-1}(\mathbf{c_{j}}-\mathbf{c_{k}}), 
\end{flalign}
where $\mathbf{\Sigma_{x}}=\sigma_{s}\mathbf{I_{2}}$,  $\mathbf{\Sigma_{c}}=\sigma_{m}\mathbf{I_{3}}$, and $\mathbf{I_{k}}$ is an identity matrix of size $k$. Furthermore, let $\mathbf{z_k}=[\mathbf{x_k},\mathbf{c_k}]^T$ and
\begin{equation}
\mathbf{\Sigma_{k}}=\begin{bmatrix}\mathbf{\Sigma_{x}}\quad\mathbf{0_{2\times3}}\\
    \mathbf{0_{3\times2}}\quad\mathbf{\mathbf{\Sigma_{c}}}
    \end{bmatrix}, \label{eq:covariance}
\end{equation}
where $\mathbf{0_{m\times n}}$ is an $m \times n$ zero matrix. Then, (\ref{eq:eucdist_general}) becomes
\begin{align}
d\left(p_j,\psi_k\right)=(\mathbf{z_{j}}-\mathbf{z_{k}})^{T}\mathbf{\Sigma_{k}}^{-1}(\mathbf{z_{j}}-\mathbf{z_{k}}).\quad\quad\label{eq:simplified}
\raisetag{1.5\normalbaselineskip}
\end{align}

We assume that for each partition $\pi$,\emph{ each pixel $p_j$ with features $\mathbf{z_{j}}$ is best described by one and only one cluster $\psi_k$ which is normally distributed with a mean (centroid) $\mathbf{z_{k}}$ and covariance $\mathbf{\Sigma_{k}}^{-1}$}. The distribution of $\mathbf{z_{j}}$ is therefore given by
\begin{equation}
\hspace{-0.5em} P(\mathbf{z_{j}}|\mathbf{z_{k}},\mathbf{\Sigma_{k}})=\frac{1}{2\pi^{5/2}|\mathbf{\Sigma_{k}}|^{1/2}}e^{\frac{-1}{2}(\mathbf{z_{j}}-\mathbf{z_{k}})^{T}\mathbf{\Sigma_{k}}^{-1}(\mathbf{z_{j}}-\mathbf{z_{k}})}.\label{eq:zjdist}
\end{equation}
The corresponding log-likelihood $l(\mathbf{z_{j}}|\mathbf{z_{k}})$ is then given by
\begin{align*}
l(\mathbf{z_{j}}|\mathbf{z_{k}}) & =-\frac{1}{2}(\mathbf{z_{j}}-\mathbf{z_{k}})^{T}\mathbf{\Sigma_{k}}^{-1}(\mathbf{z_{j}}-\mathbf{z_{k}})-\ln\left(2\pi^{5/2}|\mathbf{\Sigma_{k}}|^{1/2}\right)\label{eq:lzjzk}\nonumber\\
 & =-\frac{1}{2}d(\mathbf{z_{j}},\mathbf{z_{k}})-\alpha,\nonumber 
\end{align*}
where $d(\mathbf{z_{j}},\mathbf{z_{k}})=(\mathbf{z_{j}}-\mathbf{z_{k}})^{T}\mathbf{\Sigma_{k}}^{-1}(\mathbf{z_{j}}-\mathbf{z_{k}})$
and $\alpha=\ln\left(2\pi^{5/2}|\mathbf{\Sigma_{k}}|^{1/2}\right)$. Hence, with $\mathbf{z_{j}}\sim\mathcal{N}(\mathbf{z_{k}},\mathbf{\Sigma_{k}}^{-1})$, the distance in (\ref{eq:simplified}) is equivalent to the log-likelihood of the point $\mathbf{z_{j}}$ (without the constant offset corresponding to the normalization factor). Therefore, \emph{minimizing the distance $d\left(p_j,\psi_k\right)$ is equivalent to maximizing $l(\mathbf{z_{j}}|\mathbf{z_{k}})$}.

\subsection{Cluster assignment probability for growing}\label{sub:clusterprob}
The probability that a pixel $p_{j}$ is assigned to a cluster $\psi_{i}$ is then given by
\begin{equation}
P(p_{j}\in \psi_{i}|S)=P\left({d}(\mathbf{z_{j}},\mathbf{\bar{z}_{i}})=\min_{\psi_{k}\in\pi}{d}(\mathbf{z_{j}},\mathbf{\bar{z}_{k}})\right),\label{eq:piinpj}
\end{equation}
where $\mathbf{\bar{z}_{k}}=\mathbb{E}\left[\mathbf{z}|\psi_{k}\right]$ is the expected value of $\mathbf{z}$ within a cluster $\psi_{k}$. That is, the probability that a pixel $p_j$ is assigned to cluster $\psi_i$ is given by the probability that the distance between $\mathbf{z_j}$ and the centroid $\mathbf{\bar{z}_i}$ is the minimum distance among all the clusters centroids $\mathbf{\bar{z}_k}$. Since ${d}(\mathbf{z_{j}},\mathbf{\bar{z}_{i}})$ follows a chi-squared distribution with $n$ degrees of freedom, where $n$ is the dimensionality of $\mathbf{z}$, the cluster assignment probability is the probability that the sample ${d}(\mathbf{z_{j}},\mathbf{\bar{z}_{i}})\sim\chi_{n}^{2}$ is the minimum among the i.i.d. samples ${d}(\mathbf{z_{j}},\mathbf{\bar{z}_{k}})\sim\chi_{n}^{2},\forall \psi_k\in\pi$.

The distribution of the minimum over $\eta$ samples of a distribution with CDF $F(\cdot)$ is given by
\begin{equation}
F_{(1)}(x)=1-(1-F(x))^{\eta}\label{eq:fmin-2}.
\end{equation}
For $x\sim\chi_{n}^{2}$, 
\begin{equation}
F(x)=\frac{\gamma(n/2,x/2)}{\Gamma(n/2)},\label{eq:chiF}
\end{equation}
where $\Gamma(\cdot)$ is the gamma function and $\gamma(\cdot,\cdot)$
is the lower incomplete gamma function. Equation (\ref{eq:fmin-2}) then becomes
\begin{equation}
F_{(1)}(x)=1-\left(1-\frac{\gamma(n/2,x/2)}{\Gamma(n/2)}\right)^{\eta}\label{eq:fmin-3}.
\end{equation}

With $x=d(\mathbf{z_{j}},\mathbf{\bar{z}_{i}})$ and $n=5$, for our scenario $F_{(1)}(x)$ thus corresponds to the probability that another cluster is closer than $\psi_i$ to the pixel $p_j$. Hence, it follows that 
\begin{align}
  P(p_{j}\in \psi_{i}|S)&=1-F_{(1)}(d(\mathbf{z_{j}},\mathbf{\bar{z}_{i}}))\nonumber\\
  P(p_{j}\in \psi_{i}|S)&=\left(1-\frac{\gamma(2.5,{d}(\mathbf{z_{j}},\mathbf{\bar{z}_{i}})/2)}{1.33}\right)^{\eta},\label{eq:P_pjInpPsiI}
\end{align}
which is thus the equation that guides pixel-cluster assignments for the region growing process.

\subsection{Pixel probability estimation}
Given the set of clusters $\pi^{(m)}=\left\{ \psi_{1}^{(m)},\psi_{2}^{(m)},\ldots,\psi_{|S|}^{(m)}\right\}$ generated at the $m$-th iteration of the algorithm, the expected class likelihood $\bar{c}_i^{(m)}$ value within each cluster $\psi_i^{(m)}$ is estimated as the average of the scores $c_{j}$ associated to its pixels $p_j\in\psi_k^{(m)}$, weighted according to the probability $P(p_{j}\in \psi_{i}^{(m)}|S^{(m)})$ of pixel-cluster assignment. That is,
\begin{align}\label{eq:w_avg}
    \bar{c}_{i}^{(m)}&=P\left(\psi_i^{(m)} \in \mathcal{F}|S^{(m)}\right)  \nonumber \\
    &=\frac{\sum_{p_j\in \psi_{i}^{(m)}}c_jP(p_j \in \psi_i^{(m)}|S^{(m)})}{\sum_{p_j\in \psi_{i}^{(m)}}P(p_j \in \psi_i|S)^{(m)}}.\quad\quad
    \raisetag{1.5\normalbaselineskip}
\end{align}
Then, $\bar{c_i}^{(m)}$ is the refined class probability for all pixels $p_j\in\psi_i^{(m)}$, i.e.,
\begin{equation}
\bar{c}_j^{(m)}=P(p_j\in\mathcal{F}|\pi^{(m)})=\bar{c}_i^{(m)}.
\end{equation}
In cases where no seed is sufficiently similar to a given pixel, the probabilities of assigning this pixel to any cluster will be low and the growing process will end without any assignment for this pixel. We refer to these elements as \textit{orphan pixels}. In iterations where a pixel $p_o$ remains orphan, i.e., $p_o\notin\psi_i^{(m)},\forall\psi_i\in\pi^{(m)}$, we keep its originally predicted score $c_o$ as $\bar{c}_o^{(m)}=P(p_o\in \mathcal{F}|\pi^{(m)})$.

Let $\Pi=\{\pi^{(1)},...,\pi^{(n_s)}\}$ represent the set of all partitions generated by the multiple Monte Carlo iterations. With enough iterations, we can approximate the distribution 
\begin{equation}
    P(p_{j}\in\mathcal{F}|\pi) \approx \sum_{\pi^{(m)} \in \Pi}P(p_{j}\in\mathcal{F}|\pi^{(m)})\delta_{\pi}(\Pi),
\end{equation}
where $\delta_{\pi}(\Pi)$ is the Dirac delta function, which is equal to one if $\pi \in \Pi$ and zero otherwise. Marginalizing over the set of partitions $\Pi$, we have
\begin{align}
    \hspace{-10pt} P(p_{j}\in\mathcal{F})&=\int_{\Pi} P(p_{j}\in\mathcal{F}|\pi)P(\pi)\nonumber\\
    &\approx\frac{1}{n_{s}}\sum_{m}P(p_{j}\in\mathcal{F}|\pi^{(m)})=\frac{1}{n_{s}}\sum_{m}\bar{c_{j}}^{(m)}\label{eq:probFj},
\end{align}
such that the final refined class probability for each pixel $p_j$ is given by
$\Tilde{c_j}= P(p_{j}\in\mathcal{F})$.

\hspace{0pt}

\textbf{Variance estimation:}\label{sub:mult_prgr} In addition to the average computed in (\ref{eq:probFj}), it is also possible to compute for each pixel the variance of the estimations provided by the multiple Monte Carlo iterations. Analogously to the computation of the average $\Tilde{c_j}$, the variance $\Tilde{\sigma}_{{j}}^2$ across partitions can be computed as
\begin{equation}
    \Tilde{\sigma}_{{j}}^2=Var\left[P\left(p_j\in\mathcal{F}|\Pi\right)\right]=\frac{1}{n_s}\sum_{m}\left(\bar{c_{j}}^{(m)}-\Tilde{c_j}\right)^2. \label{eq:var}
\end{equation}
As demonstrated in Section \ref{sec:experiments}, the variance can be exploited as a measure of uncertainty that is highly correlated with segmentation accuracy. In practice, we observe that for significantly coarse predictions, it is advantageous to run the overall pRGR algorithm more than once to further improve the quality of segmentation. Let $r$ denote the ordinal index for each complete run in a set of runs $R=\{1,...,|R|\}$. Then, including the index $r$ in (\ref{eq:probFj}), each run provides an estimate $\Tilde{c_j}^{(r)}=P(p_j\in\mathcal{F}|\Pi^{(r)})$ for a pixel $p_j$. To obtain a final estimation $P(p_j\in\mathcal{F})$, we exploit inverse variance weighting to combine the estimations provided by each run. That is,
\begin{align}\label{eq:invvar}
    P(p_j\in\mathcal{F})&=\frac{\sum\limits_{r\in R}\nicefrac{\Tilde{c_j}^{(r)}}{\Tilde{\sigma}_{{j}}^2}^{(r)}}{\sum\limits_{r\in R}\nicefrac{1}{\Tilde{\sigma}_{{j}}^2}^{(r)}}.
\end{align}

\subsection{Initialization and update of cluster statistics}\label{sec:initialization}
As mentioned above, we assume clusters are normally distributed according to $\mathcal{N}(\mathbf{z_k},\mathbf{\Sigma_{k}}^{-1})$, which implies a normally distributed likelihood function. Moreover, to allow for flexible clusters that adapt to local image and prediction characteristics, similar to \cite{ban2018superpixel,uziel2019bayesSppx}, we update the terms in the spatial and color covariances in (\ref{eq:covariance}) separately, i.e.,  
\begin{align}
    \mathbf{\Sigma_{x}}=\begin{bmatrix}\sigma^2_{x} & 0\\
        0 & \sigma^2_{y}
    \end{bmatrix},\quad
    \mathbf{\Sigma_{c}}=\begin{bmatrix}
        \sigma^2_{l} & 0 & 0\\
        0 & \sigma^2_{a} & 0 \\
        0 & 0 & \sigma^2_{b}
    \end{bmatrix},
\end{align}
where $\sigma_x,\sigma_y$ are the variances along the horizontal and vertical coordinates, $\sigma_l$ is the variance of the $L$ color channel and $\sigma_{a},\sigma_{b}$ are the variances for the $a$ and $b$ channels, respectively.

\hspace{0pt}

\textbf{Initialization:} To ensure normally distributed posteriors and facilitate the update process, we initialize the mean $\mathbf{z_k}$ and covariances $\mathbf{\Sigma_{k}}$ of each cluster using conjugate prior distributions \cite{Lee2012,gelman2013bayesian}. Since the spatial and color variances are assumed to be independent, we can define Normal-inverse-chi-squared (NI$\chi^2$) prior distributions of the form 
\begin{align}
    \mu|\sigma^2 &\sim \mathcal{N}\left(\mu_0,\nicefrac{\sigma^2}{\kappa_0}\right)\nonumber\\
    \sigma^2 &\sim Inv-\chi^2\left(v_0,\sigma_0^2\right),
\end{align}
where $\mu$ and $\sigma^2$ are the means and variances for each of the five dimensions of $\left(\mathbf{z_k},\mathbf{\Sigma_{k}}\right)$, with the subscripts dropped for simplicity. The means $\mu_0$ of the normal distributions are initialized according to the locations and colors of the corresponding seeds, while $\kappa_0$ is fixed as $1$ as a seed is worth one observation of variance $\sigma^2$.

\hspace{0pt}

\noindent\textit{Spatial variances:} Initializing the inverse-chi-squared parameters $(v_o,\sigma^2_0)$ associated to the variances is more complex. Under the assumption of normally distributed clusters, the expected size of a cluster is directly proportional to the expected values of its spatial variances. Since the inter-seed spacing is known in the form of the sampled parameter $\gamma$, we expect the average cluster sizes to be proportional to $\gamma\times\gamma$. Thus, the spatial variances can be initialized as 
\begin{align}
    \sigma^2_{0,x}&=\sigma^2_{0,y}=\lambda\gamma\times\lambda\gamma,
\end{align}
where $\lambda$ is an empirically defined proportionality constant. To allow clusters to grow larger and reach lower confidence areas without nearby seeds, based on a grid search performed on a subset of $350$ randomly sampled images from the PASCAL dataset, we use the fixed value of $\lambda=27$ in all our experiments, regardless of the CNN model used to generate the segmentation masks or the dataset under consideration.

As described in \cite{Lee2012}, the $v_0$ parameters give a sense of how many observations the corresponding prior knowledge is worth. Based on this intuition, we exploit again the fact that average expected cluster sizes are proportional to $\gamma\times\gamma$, such that $v_0 \propto \gamma^2$. 
Moreover, we note that the reliability of sample variance estimations is directly proportional to the quality of the corresponding initial seed, since it defines the initial mean values. Hence, it follows that in the case of lower quality seeds more weight must be given to the prior with respect to subsequent sample variance estimates. Combining both characteristics, 
\begin{align}
    v_{0,x} = v_{0,y} &\propto \left[\frac{\gamma}{P(s_k\in I_H)}\right]^2,\label{eq:v_0x}
\end{align}
where $P(s_k\in I_H)$ corresponds to the probability that a seed is within the high-confidence region, obtained from (\ref{eq:lhcondgamma2}).

\hspace{0pt}

\noindent\textit{Color variances:} Determining an expected cluster color variance is not as straightforward. Hence, we first examined the color statistics of clusters formed using a conventional superpixel algorithm (SLIC \cite{Achanta2012}) on the same subset of the PASCAL dataset. Multiple runs with a varying numbers of superpixels and compactness values indicated variances of approximately $\sigma^2_{l}=850$ and $\sigma^2_{a}=\sigma^2_{b}=260$ to cover $99\%$ of the samples within the superpixels. Based on these observed values, we then conducted a grid-search that led to the optimal initialization values of $\sigma^2_{0,l}=1000$ and $\sigma^2_{0,a}=\sigma^2_{0,b}=300$, which are used in all our experiments. 

Since the distribution of color similarities can change from image to image, we employ a antithetic sampling variance reduction strategy \cite{mcbook} in which initial color variance values are multiplied by a value $1\pm\rho$. A value of $\rho=0.6$ was defined for all experiments after a grid search over $[0.1:0.1:0.9]$, using the same PASCAL subset described above. That is, we initialize $\sigma^2_{0,l}=1000\times\left[1\pm\rho\right]$ and $\sigma^2_{0,a}=\sigma^2_{0,b}=300\times\left[1\pm\rho\right]$. The equivalent sample size $v_{0,\{lab\}}$ for the color variances is computed using the same approach as that used for the spatial variances, which is given by (\ref{eq:v_0x}).

Finally, as explained in Sec. \ref{sec:algorithm}, in the region growing process all clusters grow from the center outwards, as the first pixels assigned are the corresponding seed neighbors, with subsequent tentative assignments of pixels neighboring the ones just assigned. In terms of sample statistics, this means initial spatial sample variances are heavily biased towards smaller values, as the first pixels assigned are the ones nearest to the corresponding cluster's centroid. To compensate for this bias, we increase the $v_0$ weight of prior variance knowledge by multiplying it with a constant, i.e., for all the experiments we set $v_0=\alpha\left[\nicefrac{\gamma}{P(s_k\in I_H)}\right]^2$. We use $\alpha=5$ for the spatial variances and, since this bias is much lower for the color statistics, we empirically set $\alpha=0.1$ for the color variances.

{\textbf{Updates:} } As detailed in \cite{Lee2012,murphy2007conjugate}, from the combination of a NI$\chi^2$ prior with the corresponding normal likelihood, the parameters of the corresponding posteriors are then given by
\begin{align}
    v_n&=v_0+n; \quad\quad \kappa_n=\kappa_0+n; \quad\quad \mu_n=\frac{\kappa_0\mu_0+n\bar{x}}{\kappa_n};\label{eq:meanupd}\raisetag{1.6\normalbaselineskip}\\
    \sigma_n^2&=\frac{1}{{v}_n}\left[{v}_0\sigma_0^2+\sum_i\left(x_i-\bar{x}\right)^2+\frac{n\kappa_0}{\kappa_0+n}\left(\mu_0-\bar{x}\right)^2\right],\quad\quad\label{eq:varupd}
    \raisetag{1.5\normalbaselineskip}
\end{align}
where $\bar{x}$ denotes the sample mean and $n$ is the total number of samples, which corresponds to the cluster size, i.e., $n=|\psi_k|$. If sample sizes are not large enough, eventual biases in the estimation of the sample variance may arise, leading to clusters with incorrect sizes. We thus apply an update strategy in which sample variance estimations are computed only after the expected cluster sizes are reached, i.e., $|\psi_k|\geq\left[\nicefrac{\gamma}{P(s_k\in I_H)}\right]^2$.

{\textbf{Posterior:} } To compute the distances and the corresponding probabilities of assigning pixels to clusters, the posterior predictive distribution is given by a $t-$student distribution with $v_n$ degrees freedom. Since for the vast majority of iterations $v_0 \geq 30$, this posterior can be approximated as normally distributed according to $\mathcal{N}(\mu_n,\sigma_n)$.

\section{Algorithm implementation}\label{sec:algorithm}
We implement pRGR by means of a main function that invokes Alg. \ref{alg:Proposed-cluster-assignment}, which summarizes the proposed region growing process that assigns pixels to clusters. First, the main script performs the non-parametric estimation of thresholds distributions and subsequent computation of seed sampling probabilities. This script then samples an initial set of seeds $S$ and invokes Alg. \ref{alg:Proposed-cluster-assignment} for region growing. 

From the image features $\mathcal{Z}$ and the corresponding set of seeds $S$ as inputs, Alg. \ref{alg:Proposed-cluster-assignment} returns an array $L$ where each pixel is mapped to its corresponding cluster by means of an index. 

\algtext*{EndWhile}
\algtext*{EndIf}
\algtext*{EndFor}
\algrenewcommand\algorithmicindent{1.25em}
\begin{algorithm}[h]
\begin{algorithmic}[1]
\Require{$\mathcal{Z}=\{z_1,...,z_{w\times h}\}$: set of 5D pixels; }
\Statex{\hspace{\algorithmicindent}\hspace{5pt}$\mathcal{S}:$ set of seeds }
\DEFINE{$L$: cluster label assigned to each pixel}
\Statex{\hspace{\algorithmicindent}\hspace{5pt} $T$: no. of times a pixel has been sampled}
\ForAll{$j=\{1,...,{w\times h}\}$}
    \State{Initialize $T[j]=0$ and $L[j]=\emptyset$}
\EndFor
\ForAll{$s_k\in S$}
    \State{Get pixel $p_j$ at $s_k$ position $(x_k,y_k)$ }
    \State{Push element $e_j=\left[j,k,1\right]$ into priority queue $Q1$}
\EndFor
\While{$(Q1 \neq \emptyset) \text{ or } (Q2 \neq \emptyset)$ }
    \State{Pop $e_j=\left[j,k,P_{jk}\right]$ from $Q1$}
    \If{$(T[j]<\kappa) \text{ and } (L[j]=\emptyset)$  }
        \State{Draw $u\sim\mathcal{U}[0,1]$} 
        \State{Increment counter $T[j]\xleftarrow{}T[j]+1$}    
        \If{$(u<P_{jk})$}
            \State{Assign $p_j$ to cluster $\psi_k$: $L[j]=k$}
            \State{Update cluster statistics with (\ref{eq:meanupd}) and (\ref{eq:varupd})}
            \ForAll{$p_n$ 8-connected to $p_j$}
                \If{$L[n]=\emptyset$}
                    \State{\hspace{-5pt}\mbox{Compute $P_{nk}=P(p_n\in\psi_k|S)$ (\ref{eq:P_pjInpPsiI})}}
                    \State{\hspace{-5pt}Push $e_n=\left[n,k,P_{nk}\right]$ into $Q1$}
                \EndIf
            \EndFor
        \ElsIf{$T[j]<\kappa$}
            \State{Push $e_j$ into recycling queue $Q2$}
        \EndIf
    \EndIf
    \If{$Q1=\emptyset$}
        \While{$Q2 \neq \emptyset$}
            \State{Pop element $e_r=\left[r,k,P_{rk}\right]$ from $Q2$}
            \State{Recompute $\hat{P}_{rk}=P(p_r\in\psi_k|S)$ (\ref{eq:P_pjInpPsiI}) }
            \State{Push element $\hat{e}_r=\left[r,k,\hat{P}_{rk}\right]$ into $Q1$}
        \EndWhile
    \EndIf
\EndWhile
\State{\textbf{return} $L$}
\end{algorithmic}

\caption{\label{alg:Proposed-cluster-assignment}Proposed cluster assignment
algorithm.}
\end{algorithm}

Let an element $e_j=\left[j,k,P_{jk}\right]$ represent a tentative assignment of a pixel $p_j$ to a cluster $\psi_k$, with the corresponding probability $P_{jk}=P(p_j \in \psi_k|S)$ (\ref{eq:P_pjInpPsiI}). For pixels sampled as seeds, elements are created with $P_{jk}$ set to $1.0$. Inspired by the SNIC \cite{Achanta2017} implementation, such tentative assignment elements are pushed into a priority queue $Q1$ that is sorted in descending order according to the assignment probabilities $P_{jk}$. Assignments occur by popping elements from $Q1$ and sampling according to the corresponding probability. Starting from the corresponding seeds, when a pixel $p_j$ is effectively assigned to a cluster $\psi_k$, all its $p_n$ $8$-connected neighbors are evaluated: if they have not been clustered yet, elements $e_{n}=[n,k,P_{nk}]$ are pushed into $Q1$ as tentative assignments of these pixels to their now neighboring cluster $\psi_k$.

With such $8$-connectivity enforcement during growing, we ensure that a pixel is visited (sampled) a maximum of $8$ times. However, since that is only an upper-bound, we opt for an implementation that ensures that each pixel will be visited at least $8$ times before being considered orphan. This is achieved through a recycling process using a recycle queue $Q2$. When an element is popped from $Q1$ but assignment does not occur, this element is pushed into a recycling queue $Q2$ if the corresponding pixel has been sampled less than $8$ times. Whenever $Q1$ is emptied, all elements in $Q2$ are updated according to the latest clusters' statistics and re-pushed into $Q1$ for processing. Using this strategy, we ensure a fixed $\eta=8$ to be used in (\ref{eq:P_pjInpPsiI}).

Therefore, the algorithm converges once all pixels have either been assigned to a cluster or visited a maximum of $8$ times. Once in possession of the corresponding mappings of pixels to clusters returned by Alg. \ref{alg:Proposed-cluster-assignment}, the main function proceeds to compute pixel probability estimations according to (\ref{eq:w_avg}-\ref{eq:invvar}).

\textbf{Gaussian filtering:}\label{sub:postconv} Since we must approximate the posterior distribution using a finite number of Monte Carlo iterations, pixels with high uncertainty might require additional refinement steps to produce accurate results. To avoid performing a large number of iterations that would impact a relatively small number of pixels, we smooth out spuriuous pixel activations using a $3\times3$ convolution with a Gaussian kernel on top of the refined scoremaps obtained using (\ref{eq:probFj}).

\section{Experiments}\label{sec:experiments}
We evaluate the performance of pRGR on: i) the $1449$ images composing the \textit{val} set of the PASCAL VOC 2012 dataset \cite{pascal2012}; and ii) selected video sequences of the DAVIS dataset \cite{Perazzi2016,Pont-Tuset2017}. While the PASCAL dataset is arguably the most widely used benchmark for semantic segmentation, its evaluation metrics disregard regions $5$ pixels-wide around the boundaries of each object. As a consequence, often clear improvements in terms of boundary adherence are not reflected in overall mean average precision ($mAP$). For that reason, we also include results using the DAVIS dataset \cite{Perazzi2016}, which is composed of high-quality video sequences with pixel-accurate ground truth segmentation for each frame.

{\textbf{Baselines:} }We compare pRGR with its antecessor RGR and also against CRF, arguably the most widely used post-processing module for semantic segmentation. We also evaluate the combination of CRF+pRGR, in which our refinement algorithm is run on top of the predictions refined using CRF. 

{\textbf{Networks:} }To assess our method for input predictions of varied quality, four different pre-trained, publicly available semantic segmentation models are considered. First, the DeepLab-COCO-LargeFOV (here DeepLab-LargeFOV for conciseness) model \cite{Chen2014_Deeplab1}, a DeepLab model using large Field-Of-View that was also used for the evaluation of RGR in  \cite{dias_rgr_2018}. We also evaluate the refinement of predictions generated by two DeepLabV2 models \cite{Chen2015_Deeplab2}, one using a VGG \cite{simonyan2014vgg} backbone and another using a ResNet backbone \cite{he2016deep}. Finally, we assess a DeepLabV3+ model \cite{Chen2017_Deeplab3} using a Xception backbone \cite{chollet2017xception}.\footnote{The first three models are available at \url{http://liangchiehchen.com/projects/DeepLab_Models.html}. The DeepLabV3+ model can be found at \url{https://github.com/tensorflow/models/tree/master/research/deeplab}} 

As summarized in Sec. \ref{sec:relworks}, these models represent different stages of recent developments in the state of the art of semantic segmentation. From their architectures, finer segmentations are expected as one moves from DeepLab to DeepLabV2 and finally DeepLabV3, both in terms of overall accuracy as well as boundary adherence. The datasets with which these models were trained also play an important role in their performance. We note that in terms of pre-training, the DeepLab-LargeFOV model exploits annotations from the MS-COCO dataset, the \textit{trainaug} subset of PASCAL VOC 2012 and, unlike the others, the \textit{val} set of PASCAL VOC 2012 in which our evaluations are performed. In contrast, both the DeepLabV2 and DeepLabV3+ models we use in our evaluation are trained only on the \textit{trainaug} subset of VOC. Of the four models, only the DeepLabV2 (VGG) is not pre-trained on COCO.

{\textbf{Parameterizations:} }Since CRF relies on a grid-search of its hyperparameters for optimal performance, as detailed below we opted for publicly available models for which the optimal CRF configurations are also provided. Regarding RGR, for all experiments, parameterization is done as reported in \cite{dias_rgr_2018}, where a different high-confidence foreground threshold value $\tau_F$ is sampled from the distribution $U (0.5,0.9)$ in each region growing iteration.

For all the cases described above, pRGR is configured to perform $20$ Monte Carlo iterations for each class scoremap. A total of $10$ different values of the seed-spacing parameter $\gamma$ are sampled from the range $[2,\gamma_{h}]$, using systematic stratified sampling. For each $\gamma$, two iterations with antithetic color configurations are run with $\rho=0.6$ as explained in Sec. \ref{sec:initialization}. According to their output strides, the different networks under consideration require distinct levels of refinement in terms of receptive field sizes. For pRGR, this corresponds to varying the upper-limit $\gamma_{h}$, as it defines the maximum expected cluster sizes. Hence, $\gamma_{h}$ is the only parameter of pRGR that is empirically adjusted on a case-by-case basis. The values selected for our experiments are listed in Table \ref{tab:params}. For all the experiments with CRF+pRGR, $\gamma_{h}$ is set to $16$.

\begin{table}[h]
\centering
\caption{Summary of pRGR configurations for each network}
\label{tab:params}
\begin{tabular}{lcccc}
\hline
 & \multicolumn{4}{c}{\textbf{DeepLab version}} \\ \cline{2-5} 
 & \textbf{LargeFOV} & \textbf{V2 (VGG)} & \textbf{V2 (ResNet)} & \textbf{V3+} \\ \hline
\textit{Double refinement} & \checkmark & \checkmark &  &  \\
\textit{$\gamma_{h}$} & 48 & 32 & 24 & 16 \\ \hline
\end{tabular}
\end{table}

\begin{figure*}[t]
    \centering
    \begin{tabularx}{\textwidth}{YYYYYY}
    {Image} & {CNN} & {+RGR} & {+pRGR} & {+CRF} & {+CRF+pRGR} \\
    \end{tabularx}
    \includegraphics[trim={0 0cm 0 0},clip,width=\textwidth]{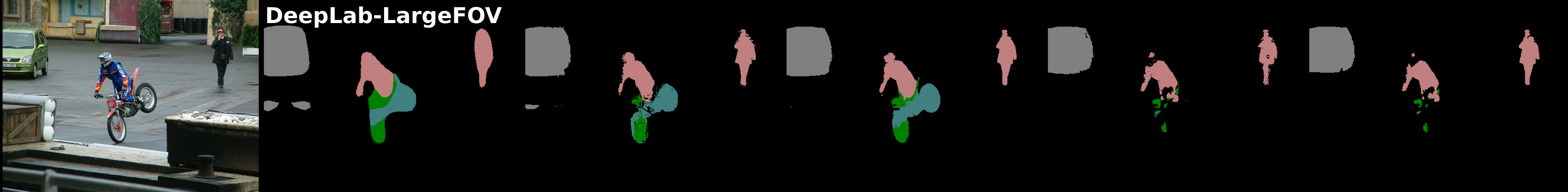}
    \includegraphics[trim={0 0cm 0 0cm},clip,width=\textwidth]{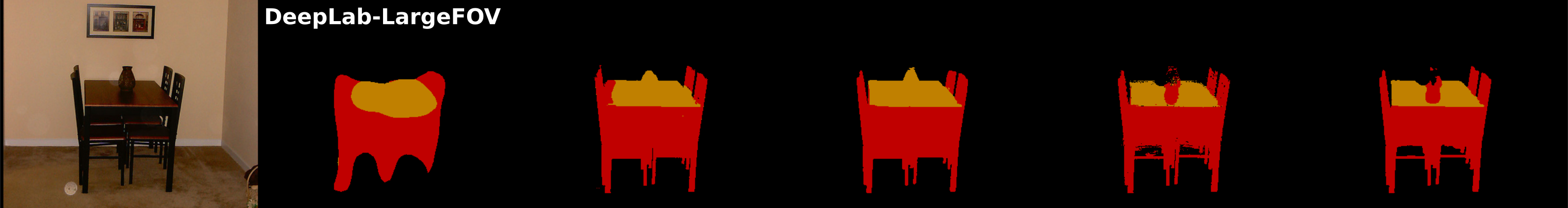}
    \includegraphics[width=\textwidth]{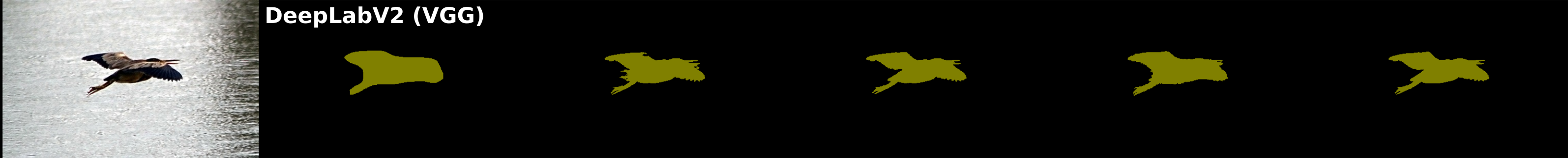}
    \includegraphics[width=\textwidth]{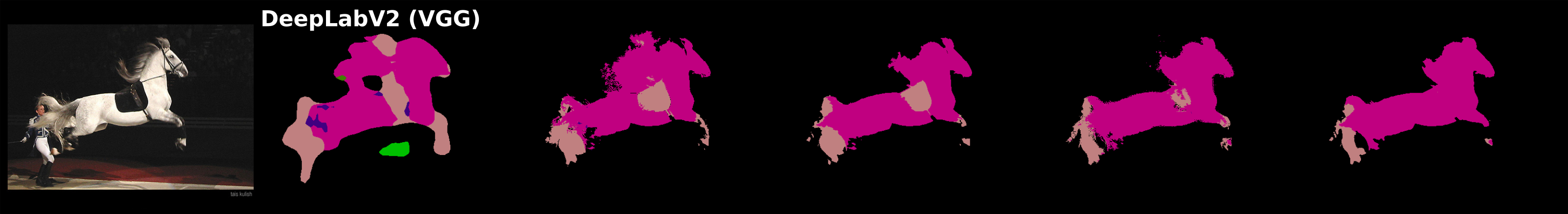}
    \includegraphics[width=\textwidth]{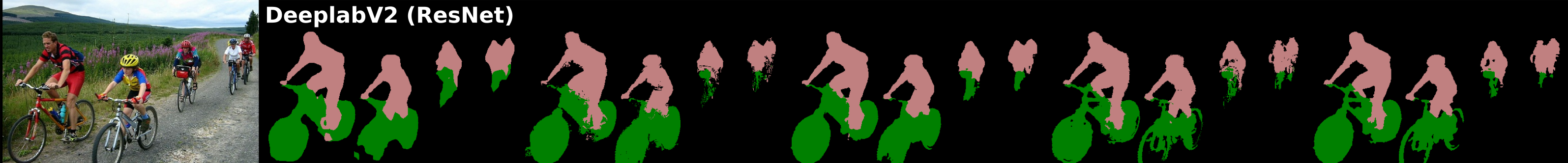}
    \caption{Qualitative results on PASCAL \textit{val} images. In the second column, overlaid names correspond to the CNNs used for each prediction. }
    \label{fig:pascalmosaic}
\end{figure*}

As illustrated in Fig. \ref{fig:pascalmosaic}, the segmentations provided by DeepLab-LargeFOV and DeepLabV2 (VGG) are fairly coarse, such that for these cases we perform two pRGR refinement steps using inverse variance weighing to combine the estimation results of each step, as explained in (\ref{eq:invvar}).

\newcolumntype{Y}{>{\centering\arraybackslash}X}

\subsection{Comparison with baselines on PASCAL}
\begin{figure}[h!]
    \centering
    \includegraphics[width=\linewidth]{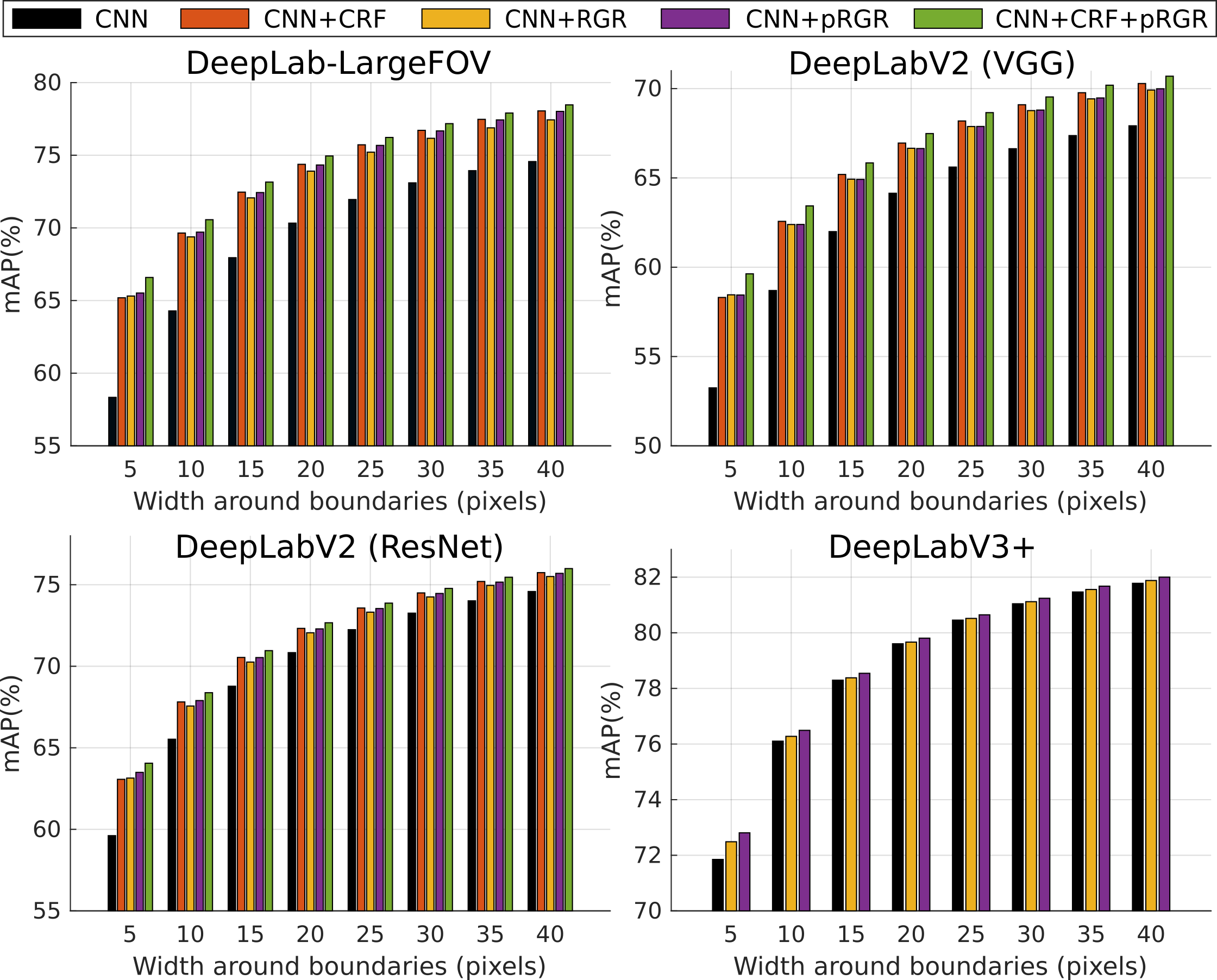}
    \caption{Summary of $mAP$ on PASCAL for regions of varying width near the object boundaries. Each color corresponds to a combination of the corresponding CNN with a refinement method, according to the legend above the figures. }
    \label{fig:vocTrimaps}
\end{figure}

\begin{figure}[h]
    \centering
        \includegraphics[width=\linewidth]{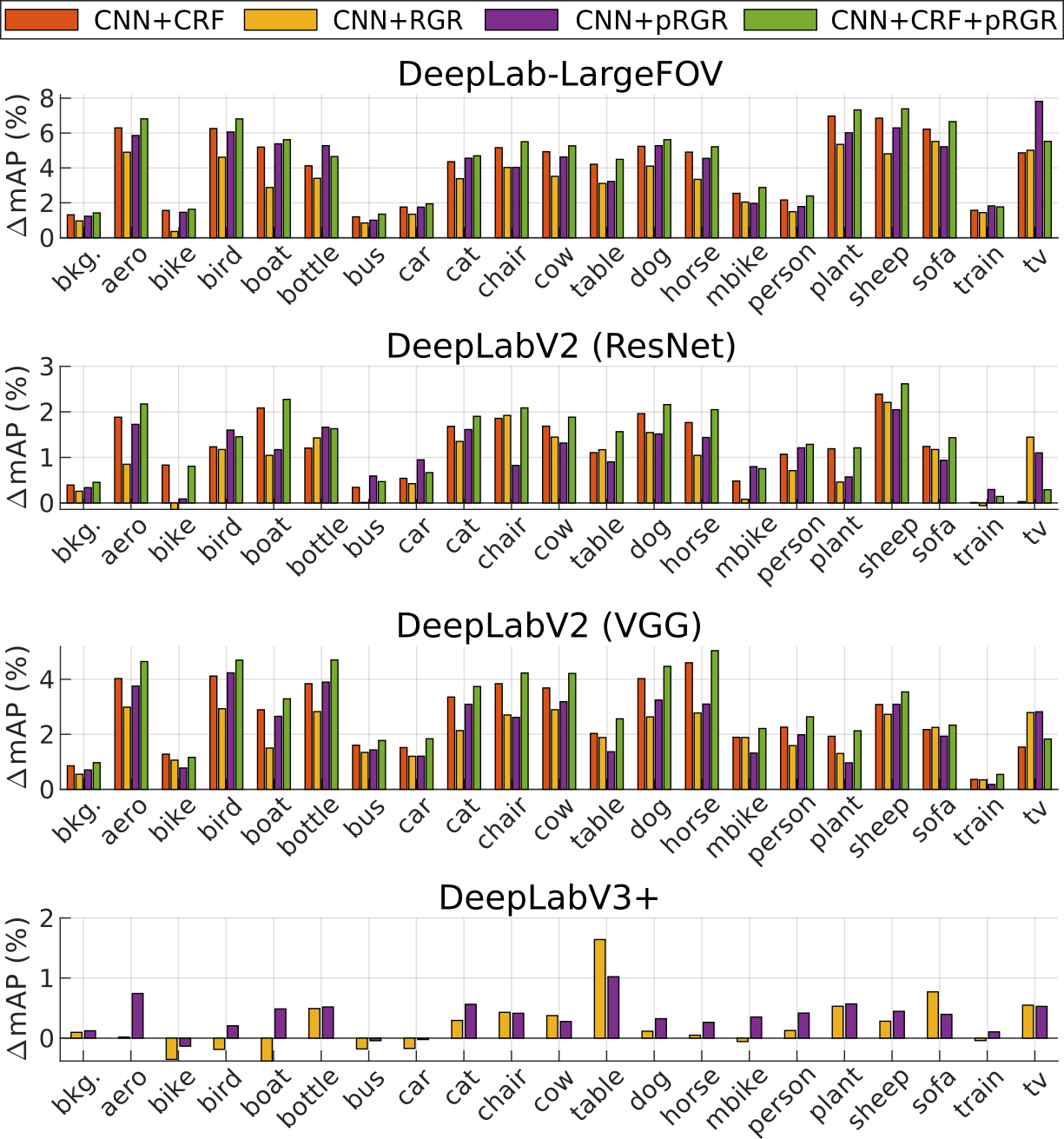}
    \caption{Improvements on segmentation accuracy ($\Delta mAP(\%)$) provided by each refinement method according to specific categories on PASCAL dataset.}
    \label{fig:voc_categ}
\end{figure}

Table \ref{tab:deeplab12} summarizes the quantitative results provided by each combination of refinement methods with the corresponding four variations of semantic segmentation networks. Since boundaries constitute a small fraction of the total image pixels, to better quantify boundary adherence we follow the strategy presented in \cite{Ghiasi2016laplacian} and also evaluate segmentation accuracy on narrower regions closer to the boundaries. Fig. \ref{fig:pascalmosaic} presents qualitative examples of segmentation masks provided by each combination of methods, while Fig. \ref{fig:vocTrimaps} shows the $mAP$ values obtained by each method as a function of the object boundary width considered in the evaluation. Finally, Fig. \ref{fig:voc_categ} details the performances of each method according to each category of the PASCAL dataset.

\begin{table}[h]
 \setlength{\tabcolsep}{12pt}
  \caption{Comparison and combination of pRGR and baselines on PASCAL dataset.}
    \label{tab:deeplab12}
    \setlength{\tabcolsep}{4pt}
        \begin{tabular}{lccccc}
            \hline
             & \multicolumn{5}{c}{\textbf{VOC 2012 - mAP(\%)}} \\ \hline
            \textbf{CNN} & \textit{No ref.} & \textit{+CRF} & \textit{+RGR} & \textit{+pRGR} & \textit{+CRF+pRGR} \\ \hline
            DeepLab-LargeFOV & 76.05 & 80.23 & 79.21 & 80.11 & 80.58 \\
            DeepLabV2 (VGG) & 68.96 & 71.57 & 70.97 & 71.22 & 71.94 \\
            DeepLabV2 (ResNet) & 76.46 & 77.65 & 77.39 & 77.54 & 77.86 \\ \hline
        \end{tabular}
    \hspace{0.25cm}
\end{table}

{\textbf{Boundary adherence:}} The results in Fig. \ref{fig:vocTrimaps} highlight how all the methods under consideration improve segmentation accuracy especially in regions near boundaries. In comparison with the results shown in Table \ref{tab:deeplab12}, even for scenarios such as DeepLabV2 (ResNet), where overall $mAP$ improvements are slightly above $+1.0\%$, the segmentation accuracy in regions $\leq 5px$ near the boundaries is improved by approximately $+3.5\%$ using pRGR. 

{\textbf{RGR vs pRGR:}} Overall, our results demonstrate that pRGR consistently outperforms RGR in all the scenarios under consideration. In comparison with its precursor RGR, the probabilistic formulation of pRGR combined with refinement iterations at different receptive field sizes reduces the occurrence of noisy predictions and minimizes the impact of false positives. This is illustrated in Fig. \ref{fig:pascalmosaic} near the bird's wings and beak, and also near the horse's crest.

{\textbf{CRF vs pRGR:}} In terms of overall accuracy, pRGR provides $mAP$ values slightly lower than the ones obtained with CRF. However, the results summarized in Fig. \ref{fig:vocTrimaps} indicate that predictions refined using pRGR are slightly better (FOV: $+0.33\%$, VGG: $+0.14\%$, ResNet: $+0.43\%$) than the ones using CRF for regions $\leq 5px$ near the boundaries. This is also exemplified near the bird's wings in Fig. \ref{fig:pascalmosaic}. On the other hand, results detailed in Fig. \ref{fig:voc_categ} for categories such as bicycles and chairs suggest that the main failure case of pRGR corresponds to enclosed regions with high amounts of false-positives, such as the internal areas of bicycles' wheels and chairs' spindles. Qualitatively, this is illustrated in the last example of Fig. \ref{fig:pascalmosaic}. As the region growing procedure is based on $8-$connectivity, it cannot correct such enclosed regions containing high amounts of false positives. In contrast, CRF is able to recover from such mistakes, which is reflected in the overall higher $mAP$ values. However, it is important to note again that pRGR is entirely unsupervised, whereas CRF must be fine-tuned to the dataset and segmentation network under consideration.

{\textbf{CRF+pRGR:}}  Our analysis suggests that, while CRF and pRGR provide similar overall performances, they have different success/failure cases. As such, combining CRF and pRGR is a potential strategy for further refining segmentation masks, which is corroborated by the results reported as $CRF+pRGR$ in Table \ref{tab:deeplab12} and Figs. \ref{fig:pascalmosaic} and \ref{fig:vocTrimaps}. In all the evaluated scenarios, this combination significantly outperforms CRF alone, especially in regions near boundaries as shown quantitatively in Fig. \ref{fig:vocTrimaps} and can be noticed in the chairs' and bird's details in Fig. \ref{fig:pascalmosaic}. Moreover, the fourth example in Fig. \ref{fig:pascalmosaic} illustrates how pRGR can also mitigate some false positives partially attenuated by CRF, such as misdetections near the saddle and the horse's knee. Finally, results combining $CRF+pRGR$ also demonstrate that, if the amount of false-positives is reduced and enough high-quality seeds are available, pRGR can also improve segmentations in the failure case scenarios. 

\subsection{Refinement of DeepLabV3+ predictions}
Table \ref{tab:deeplab3} summarizes the performances of DeepLabV3+ before and after refinement using RGR and pRGR, for experiments on the PASCAL and DAVIS datasets. Unlike the previous experiments, here the CRF baseline is not considered, since no CRF implementation optimized for DeepLabV3+ is currently available.

\begin{table}[h]
    \centering
        \caption{Effect of RGR and pRGR for refinement of DeepLabV3+ predictions.}
        \begin{tabular}{lllccc}
        \hline
        \textbf{CNN} & \textbf{Dataset} &  & \textit{No ref.} & \textit{+RGR} & \textit{+pRGR }\\ \hline
        \multirow{3}{*}{DeepLabV3+} &\textbf{VOC 2012} & \textit{mAP(\%)} & 82.20 & 82.41 & 82.56 \\ \cline{2-6}
        & \multirow{2}{*}{\textbf{DAVIS 2017}} & \textit{$\mathcal{J}$ mean} & 76.29 & 80.19 & 80.47 \\
        & & \textit{$\mathcal{F}$ mean} & 76.41 & 80.10 & 80.30 \\ \hline
        \end{tabular}
    \label{tab:deeplab3}
\end{table}

\begin{figure*}[t]
    \centering
    \includegraphics[width=\linewidth]{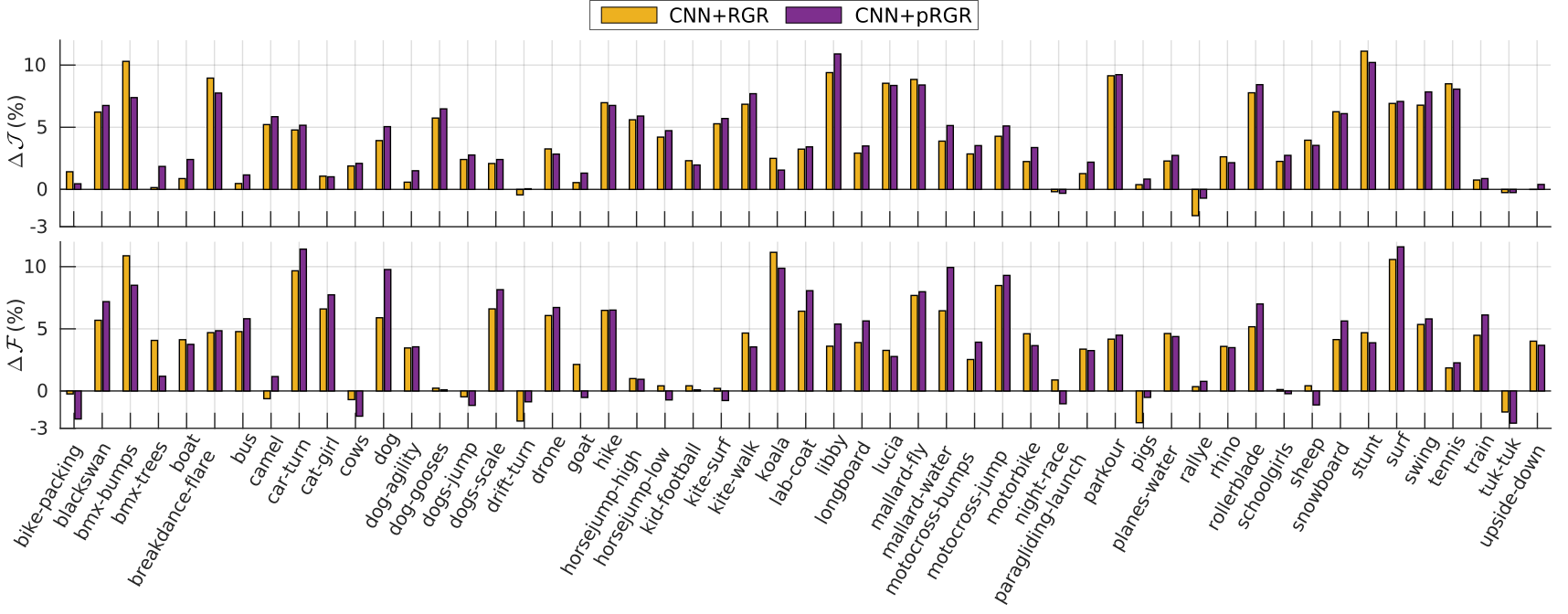}
    \caption{Improvements on segmentation accuracy provided by each refinement method according to specific sequences of the DAVIS dataset. \textit{Top:} variations in $\mathcal{J} mean$. \textit{Bottom:} variations in $\mathcal{F} mean$}
    \label{fig:DAVIS_categ}
\end{figure*}

From Table \ref{tab:deeplab3} and the results in the lower right corner of Fig. \ref{fig:vocTrimaps}, experiments on the PASCAL dataset using DeepLabV3+ once again indicate that, although the gains in overall $mAP$ are relatively small ($\approx0.36\%$), both RGR and pRGR provide non-negligible improvements in terms of boundary adherence even for state-of-the-art semantic segmentation networks ($\approx1.0\%$ for regions $\leq 5px$ near boundaries). 

To further validate this observation, we selected $53$ video sequences listed in Fig. \ref{fig:DAVIS_categ} from the DAVIS 2016 \cite{Perazzi2016} and 2017 \cite{Pont-Tuset2017} datasets for further experimentation with the same DeepLabV3+ model. Since this model is trained for the $21$ PASCAL categories, we selected only sequences where the target objects are within this set of categories.

As previously mentioned, the DAVIS evaluation metrics include both overall intersection-over-union (or Jaccard-index) $\mathcal{J}$ and also a contour accuracy metric $\mathcal{F}$ that assesses specifically the accuracy near object boundaries. Table \ref{tab:deeplab3} contains the results obtained using both metrics for predictions before and after RGR and pRGR refinement. Since the DAVIS annotations take into consideration all the pixels composing object boundaries, in this dataset, improvements in terms of boundary adherence have higher impact in final performance metrics than the ones observed for experiments on the PASCAL dataset. The results reveal improvements in the order of $\approx4.0\%$ by both refinement methods, with pRGR marginally yet consistently outperforming its antecessor in both metrics. 

\begin{figure}[h]
    \centering
    \begin{tabularx}{\linewidth}{YYY}
    {Image} & {DeepLabV3+} & {+RGR}  \\
    \end{tabularx}
    \includegraphics[trim={0 0cm 0 0},clip,width=\linewidth]{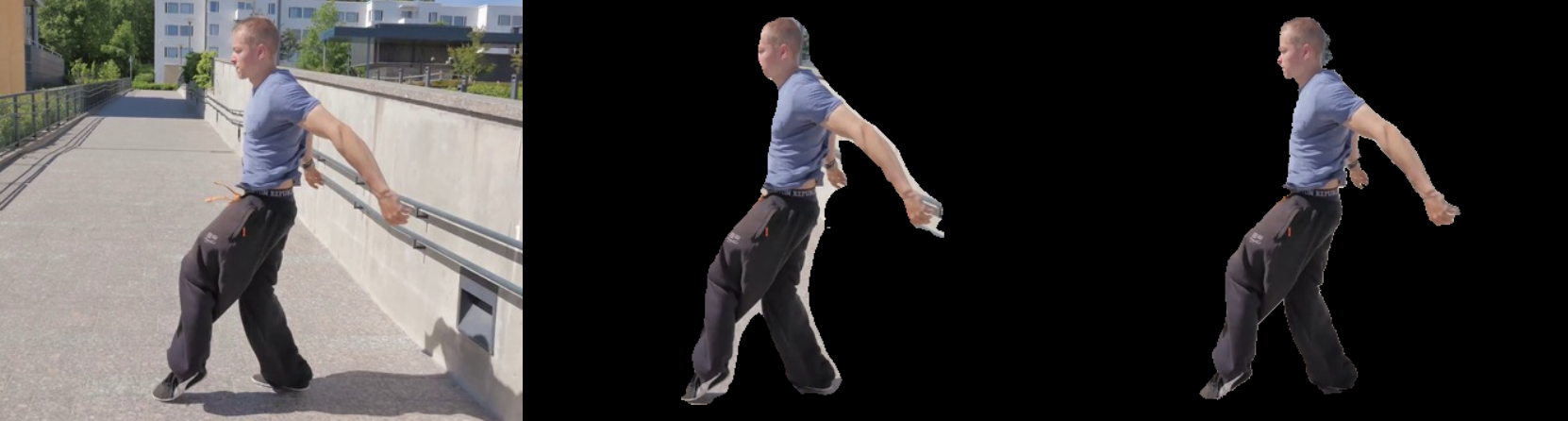}
    \includegraphics[trim={0 0cm 0 0},clip,width=\linewidth]{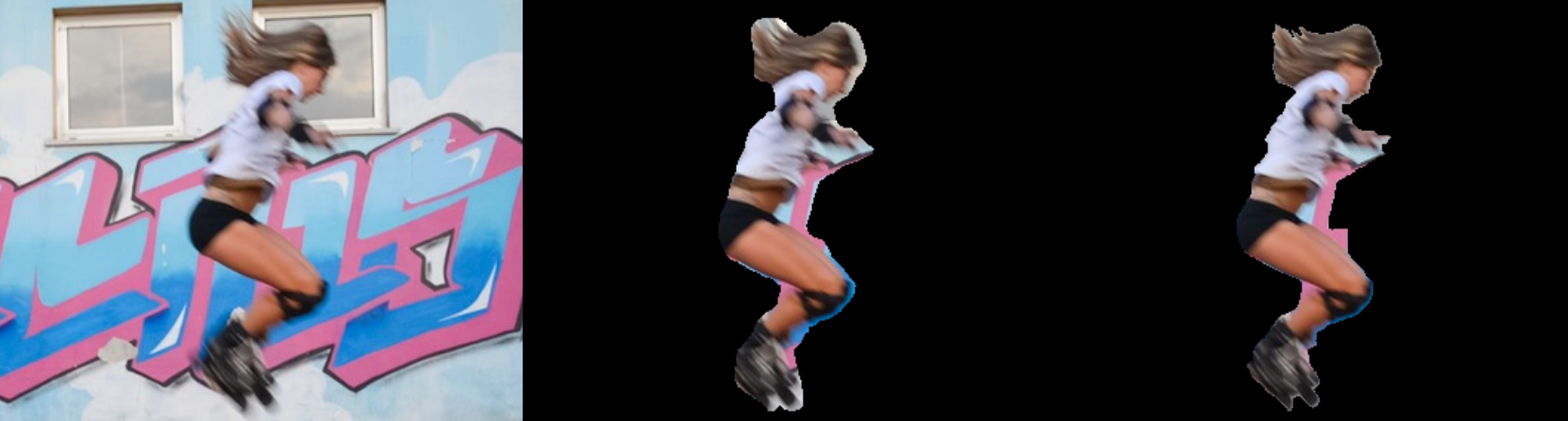}
    \includegraphics[trim={0 0cm 0 0},clip,width=\linewidth]{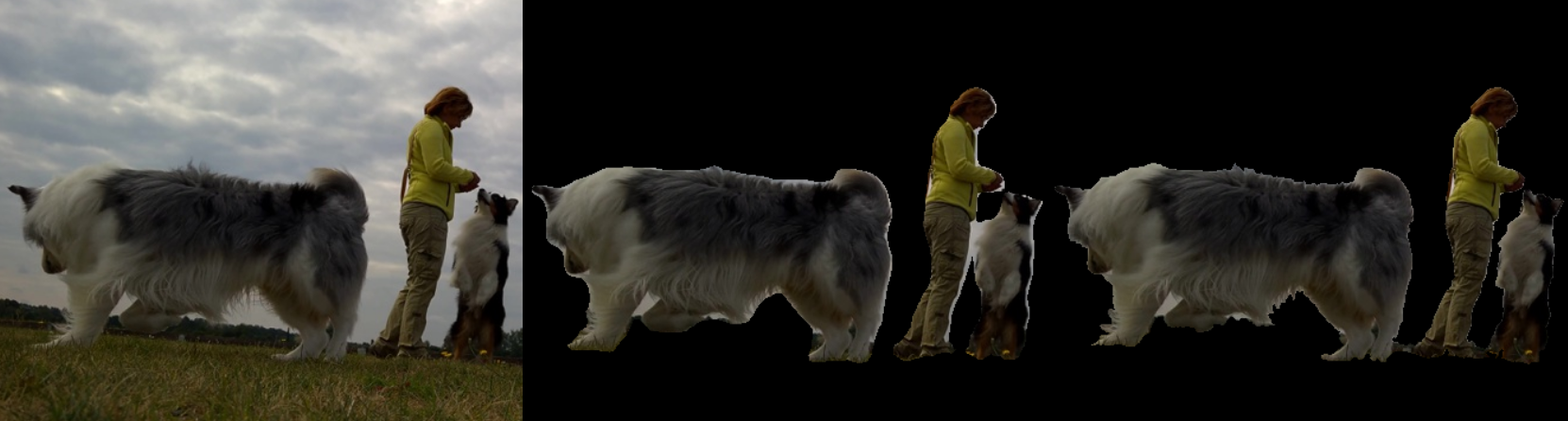}
    \caption{Examples of details recovered through pRGR refinement of DeepLabV3+ predictions for images in the DAVIS dataset.}
    \label{fig:davismosaic}
\end{figure}

Results for the $\mathcal{F}$ metric demonstrate that pRGR provides large improvements in terms of boundary adherence, with a $3.9\%$ increase in mean $\mathcal{F}$. Fig. \ref{fig:davismosaic} shows qualitative examples of such improvements. In all the examples, we observe how the refined segmentation masks include fewer pixels comprising the surrounding background. In the first two images, details such as the people's hair and feet are recovered. In the last image, the refined segmentation properly adheres to the dogs' fur, and correctly separates the person from the dog.

From the results for the individual DAVIS sequence detailed in Fig. \ref{fig:DAVIS_categ}, lower performances are observed for some sequences containing vehicles and animals as targets. For the first case, failures mostly arise from propagating false-positive detections of shadows under vehicles. In the case of animals, limb extremities can be lost when such elongated structures are detected with low confidence, are far from the animal's body, and share a similar color with the surrounding background. Yet, we emphasize that significant improvements are observed for most scenarios evaluated.

\subsection{Uncertainty estimation}
As noted by Kendall \& Gal \cite{kendall2017}, the normalized scores provided by CNNs do not necessarily reflect the uncertainties of these classification models. In \cite{mukhoti2018}, Bayesian Deep Learning is exploited using Monte Carlo dropout and Concrete dropout to capture uncertainties of a DeepLabV3+ model for semantic segmentation. In our pRGR framework, the variance of estimations across multiple Monte Carlo refinement iterations (computed using (\ref{eq:var})) can be exploited as a measure of classification uncertainty. To validate this claim, we evaluated the $mAP$ values on the PASCAL dataset for increasingly high thresholds of variance values. Similarly, we establish a baseline for comparison by computing the accuracy of the original network's predictions for increasingly high thresholds of predicted class scores. 

\begin{figure}[h]
    \centering
    \includegraphics[width=\linewidth]{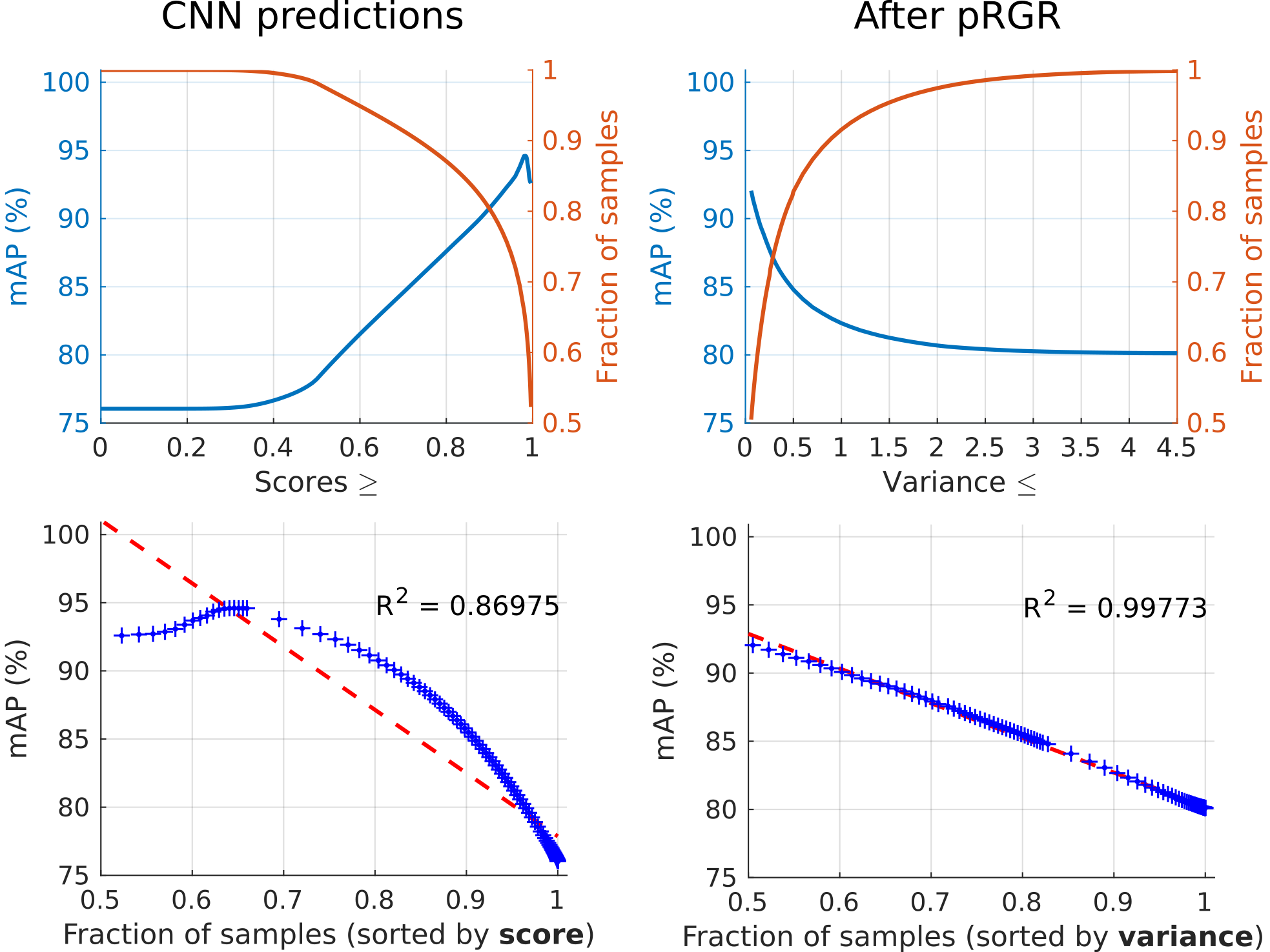}
    \caption{Correlation between segmentation accuracy and \textit{left)} original CNN prediction scores; \textit{right)} variance across pRGR Monte Carlo refinement iterations. }
    \label{fig:unc_corr}
\end{figure}

Fig. \ref{fig:unc_corr} presents the results collected for experiments using DeepLab-LargeFOV predictions. For both cases, the curves in the top row suggest a significant correlation between prediction scores (for CNN predictions) and estimated variances (from pRGR outputs) with the actual segmentation accuracy. %

However, for the CNN predictions, sharper slope variations are observed both at the beginning and the end of the $mAP$'s curve. Since for both cases the fraction of samples covered varies non-linearly as the threshold values increase, we also analyze the accuracy vs. the fraction of samples to assess the correlation between segmentation quality and uncertainty estimations. More specifically, the graphs in the bottom row of Fig. \ref{fig:unc_corr} are obtained by plotting the left y-axis vs. the right y-axis for each corresponding plot from the top row. This analysis corresponds to assessing how segmentation accuracy decays as larger fractions of samples with increasingly high uncertainty are considered.

This analysis clearly demonstrates a linear relationship between pRGR's variance estimations and segmentation uncertainty. The plots on the right column indicate that the correlation between pRGR's estimated variance and final segmentation accuracy are very strong, which is numerically confirmed by a correlation coefficient of $R^2\geq0.99$. For the sake of brevity, we only provide the plots using the DeepLab-LargeFOV, but coefficients $R^2\geq0.99$ are also observed for the DeepLabV2 (VGG), DeepLabV2 (ResNet) and DeepLabV3+ network configurations.

\section{Conclusion}\label{sec:conclusions}
We have presented pRGR, an updated version of our fully unsupervised RGR algorithm for semantic segmentation refinement. By combining concepts of probability theory, Bayesian estimation, and variance reduction, pRGR not only provides a solid mathematical foundation for RGR, but also further improves the quality of the segmentations obtained after refinement. 

Through a Monte Carlo formulation where seed-spacing parameters are sampled in a stratified manner, pRGR evaluates varied receptive field sizes across its multiple region growing iterations of high-confidence seeds. Combined with a strategy where cluster covariances are initialized using conjugate priors and updated as pixel-cluster assignments occur, these new features allow pRGR to refine segmentation masks to significantly higher pixel-accuracy levels. As demonstrated through experiments on the PASCAL and DAVIS datasets using four different configurations from the DeepLab family, segmentation predictions refined with pRGR are improved especially in terms of boundary adherence and removal of false-positive pixel labels.

Moreover, the practical relevance of the proposed algorithm also includes a possible combination with the DenseCRF model to further improve the segmentation quality provided by each of these methods alone, as demonstrated by our experimental results. Finally, thanks to its Monte Carlo framework for estimation, pRGR also generates variance estimates that show a strong inverse correlation with the final segmentation accuracy values. In other words, pRGR variance values can be exploited for uncertainty estimation of segmentation predictions, which expands its range of applications to scenarios such as active learning \cite{gal2017deep}, human-in-the-loop systems for image labeling \cite{maninis2018deep}, and semi- or weakly-supervised methods for image segmentation \cite{kolesnikov2016seed,song2018seednet}.

%
%
%
\IEEEtriggeratref{45}
\bibliographystyle{IEEEtran}
\bibliography{IEEEabrv,refs}

\begin{IEEEbiography}[{\includegraphics[width=1in,height=1.25in,clip,keepaspectratio]{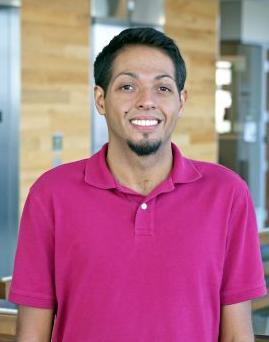}}]{Philipe Dias}
is a PhD student in Electrical and Computer Engineering at Marquette University.
As result of a Double Masters Degree, he received his M.Sc. in Information Technology from the Hochschule Mannheim (Germany) and his Master in Electrical \& Computer Eng. from the Federal University of Technology (UTFPR, Brazil). His research topics include supervised and unsupervised learning, combined with probability theory and stochastic simulation for applications in agriculture, semantic segmentation and image annotation. His PhD studies included a period at the University of Genoa (Italy), as part of a collaboration where Philipe has been applying such techniques for image segmentation and gaze estimation in healthcare-related scenarios.
\end{IEEEbiography}

\begin{IEEEbiography}[{\includegraphics[width=1in,height=1.25in,clip,keepaspectratio]{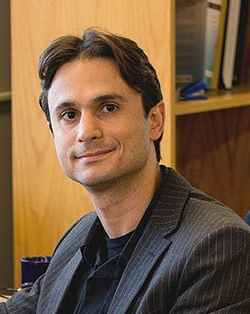}}]{Henry Medeiros}
is an Assistant Professor of Electrical and Computer Engineering at Marquette University. His research interests include computer vision and robotics with applications in areas such as manufacturing, agricultural automation, and public safety. He has published over forty journal and peer-reviewed conference papers and holds several US and international patents. Before joining Marquette, he was a Research Scientist at the School of Electrical and Computer Engineering at Purdue University and the Chief Technology Officer of Spensa Technologies, a technology start-up company located at the Purdue Research Park. He is a senior member of the IEEE and has been an associate editor for the IEEE International Conference on Robotics and Automation and the IEEE International Conference on Intelligent Robots. He received his Ph.D. from the School of Electrical and Computer Engineering at Purdue University as a Fulbright scholar. \end{IEEEbiography}

\end{document}